\documentclass[conference]{IEEEtran}
\usepackage{booktabs}
\usepackage{float}
\usepackage{graphicx}
\usepackage{dsfont}
\usepackage{amsmath}
\usepackage[hypertexnames=false]{hyperref}
\usepackage{algpseudocode}
\usepackage{algorithm}
\usepackage{multirow}
\usepackage{color}
\usepackage{soul}

\ifCLASSINFOpdf
\else
\fi

\begin{document}
\title{Benchmarking Pretrained Vision Embeddings for Near- and Duplicate Detection in Medical Images}

\author{\IEEEauthorblockN{Tuan Truong}
\IEEEauthorblockA{Bayer AG, Berlin, Germany\\
tuan.truong@bayer.com}
\and
\IEEEauthorblockN{Farnaz Khun Jush}
\IEEEauthorblockA{Bayer AG, Berlin, Germany\\
farnaz.khunjush@bayer.com}
\and
\IEEEauthorblockN{Matthias Lenga}
\IEEEauthorblockA{Bayer AG, Berlin, Germany\\
matthias.lenga@bayer.com}}

\maketitle

\begin{abstract}
Near- and duplicate image detection is a critical concern in the field of medical imaging. Medical datasets often contain similar or duplicate images from various sources, which can lead to significant performance issues and evaluation biases, especially in machine learning tasks due to data leakage between training and testing subsets. In this paper, we present an approach for identifying near- and duplicate 3D medical images leveraging publicly available 2D computer vision embeddings. 
We assessed our approach by comparing embeddings extracted from two state-of-the-art self-supervised pretrained models and two different vector index structures for similarity retrieval. We generate an experimental benchmark based on the publicly available Medical Segmentation Decathlon dataset. The proposed method yields promising results for near- and duplicate image detection achieving a mean sensitivity and specificity of 0.9645 and 0.8559, respectively. 
\end{abstract}


% \fi
\IEEEpeerreviewmaketitle

\section{Introduction}
Near- and duplicate image detection aims to find similar or identical images in a large data corpus and has many applications in web image retrieval and forensic tasks. Most image (near-) duplicate detection algorithms rely on matching images based on distance measures of associated image embeddings. In order to classify (near-) duplicates and non-duplicates typically a distance threshold is used. The most common handcrafted embeddings are SIFT \cite{lowe2004distinctive} and SURF \cite{bay2006surf} features. Recent works leverage Deep Learning features \cite{zhou_near-duplicate_2020, morra_benchmarking_2019, koker_identification_2021} or combine with handcrafted features such as SIFT \cite{zhou_integrating_2020} to construct features at local and global level.

In medical imaging, identical or slightly preprocessed images can be sold by different data brokers or even by the same vendor for different tasks. This is in particular risky when machine learning approaches are used. The model performance may be negatively impacted by (near-) duplicates or even evaluation biases might occur due to data leakage between train and test sets as observed in \cite{barz_we_2020}. 
Developing (near-) duplicate detection algorithms for medical images comes with several challenges. First, most medical images are acquired in 3D and need to be projected to a low-dimensional representation for similarity search. While there are plenty of powerful pre-trained 2D image feature extractors available \cite{he2016deep, dosovitskiy2020image}, there is much less work related to 3D feature extraction. Naively matching slice-wise embeddings is not trivial. For example, border slices often contain uninformative background voxels
which may cause mismatches with border slices from other
cases regardless of anatomy or modality.  This has direct implications for robustly calibrating the similarity threshold for duplicate identification.
In our approach, we aim to leverage 2D feature extractors pretrained on natural images and propose a count-based matching scheme for 3D image volumes that supports finding an optimal threshold for detecting (near-) duplicates and non-duplicates. Our main contributions are as follows:
\begin{itemize}
    \item We introduce a matching approach for 3D image volumes that relies on the majority counts from slice-wise retrievals, weakening the impacts of background slices on volume retrieval. We show how to identify an optimal threshold for both near-duplicate and duplicate image detection.
    \item We synthesize near-duplicates using various image transformations with different distortion levels. The robustness of different pretrained feature embeddings towards these transformations is quantitatively benchmarked.
    \item We evaluate the detection performance using embeddings extracted by DINO \cite{caron2021emerging, oquab2023dinov2}, the state-of-the-art self-supervised models designed for natural images, demonstrating the transferability of such models to medical imaging tasks without fine-tuning as studied in \cite{truong2021transferable}.
\end{itemize}

\section{Methods}
\begin{figure}[htp]
    \centering
    \includegraphics[width=\linewidth]{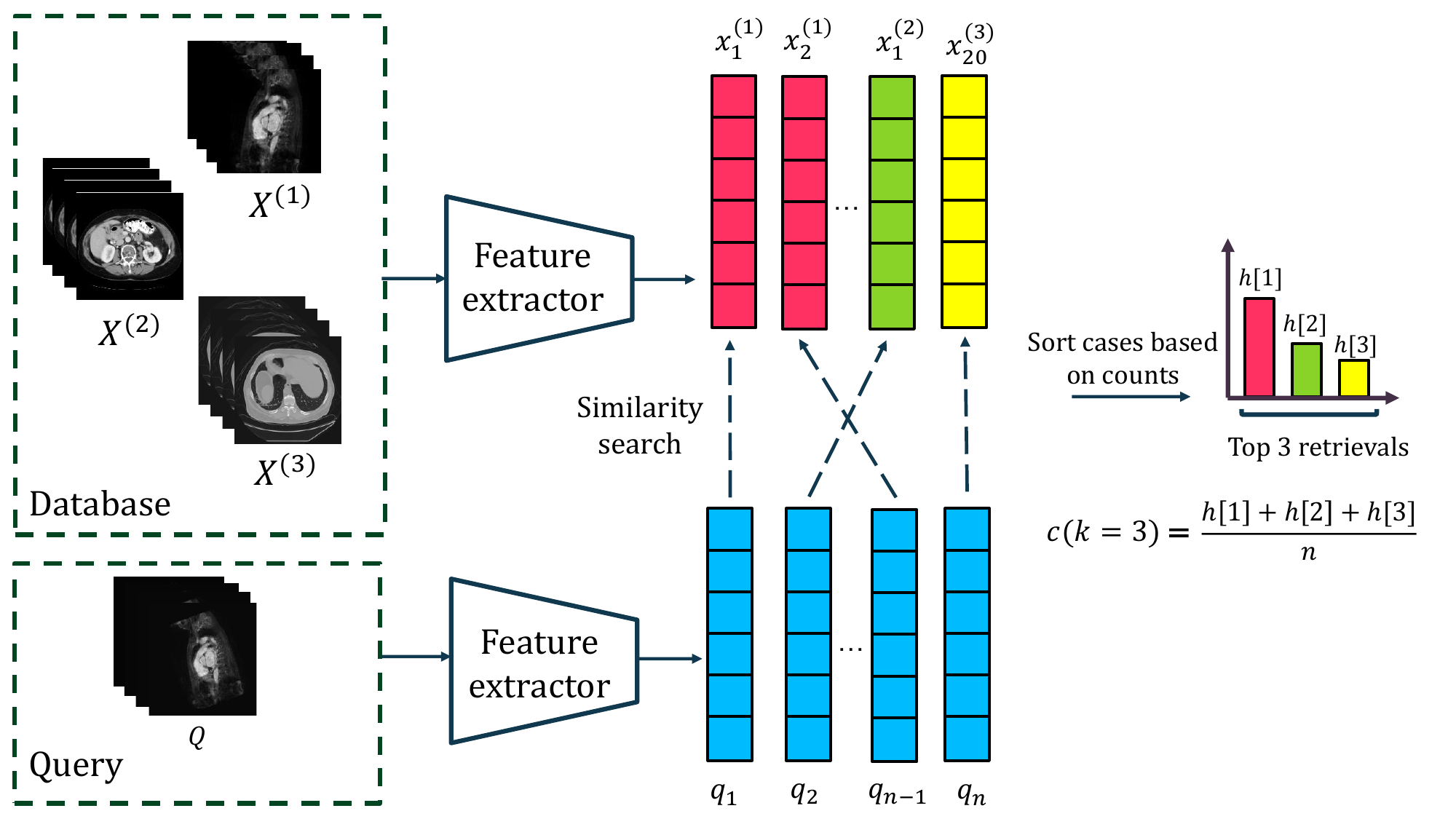}
    \caption{Retrieval at case level based on count accumulation}
    \label{fig:method}
\end{figure}

Our proposed method relies on retrieving 3D medical images based on slice-wise similarity scores and cross-volume aggregation.
As shown in Figure \ref{fig:method}, a 2D feature extractor $\phi$ is used to embed the 2D slices into a vector representation. Let $X^{(i)}$ be the image volume related to case $i$ in the database and $x^{(i)}_j := \phi(X^{(i)}_j)$ be the embedding of the $j$-th slice. In our implementation all embeddings are stored in a vector database using the FAISS backend \cite{johnson2019billion}.
% 
% Embeddings of images in the database are indexed to enable effective similarity searching. In our experiment, we choose LSH and HNSW as alternatives to benchmark performance metrics. 
% 
Let $Q$ represent a query image volume consisting of $n$ slices, and $q_m := \phi(Q_m)$ be the embedding of the $m$-th query volume slice. For all query slices $q_m (m = 1,...,n)$ we retrieve the embedding from the database which has minimal Euclidean distance with $q_m$, and denote the corresponding case with $N[q_m]$.
A histogram is built for all $n$ query slices to identify the case ID in the database with the most hits:
\begin{equation}
    h[i] = \sum_{m=1}^{n}\mathds{I}(N[q_m], X^{(i)})
\end{equation}

Here, $\mathds{I}$ is the identity function that returns 1 if $N[q_m]$ matches $X^{(i)}$ and 0 otherwise. 
The normalized count for top-$k$ predictions is then defined as
\begin{equation}
    c(k) = \frac{\sum_{l=1}^{k}h[i_l]}{n}
\end{equation}
where $i_1, ..., i_k$ denote the indices of the $k$ largest hits, i.e. $h[i_1] \geq ... \geq h[i_k] \geq h[j]$ for all $j\not\in\{i_1, ..., i_k\}$.
The normalized count has a range between 0 and 1. Based on $c(k)$ and a threshold, (near-) duplicates and non-duplicates will be identified. We conduct experiments to finetune this threshold for optimal detection performance and propose in Algorithm \ref{alg:cutoff} a method to select an optimal threshold $t_{\textrm{opt}}$ across different query sets $\mathcal{Q} := \left\{s_1, s_2, ..., s_N\right\}$ which are defined in Section \ref{results-cutoff-selection}. First, we generate a candidate list $\mathbf{T}$ of thresholds based on the maximal Youden's index associated with the Receiver Operating Characteristic (ROC) curve of each query set $s \in \mathcal{Q}$. The final optimal threshold $t_{\textrm{opt}}$ is then chosen to maximize the sum of average sensitivity and specificity across all query sets.

\begin{algorithm}
\caption{Finding optimal decision threshold}
\label{alg:cutoff}
\begin{algorithmic}[1]
\State $\mathcal{Q} := \left\{s_1, s_2, ..., s_N\right\}$
\State \textbf{init array} $\mathbf{T}$, $\mathbf{SE}$, $\mathbf{SP}$
\For {$u=1, 2, ..., N$}
\State $R_u:=$ ROC curve for $s_u$
\State $\mathbf{T}[u] \gets t_{ROC}:=\underset{r \in R_u}{\textrm{argmax}}(\textrm{sens}[r] + \textrm{spec}[r])$
\For {$v=1, 2, ..., N$}
\State $\mathbf{SE}[u, v] \gets$ sensitivity of $s_v$ at threshold $\mathbf{T}[u]$
\State $\mathbf{SP}[u, v] \gets$ specificity of $s_v$ at threshold $\mathbf{T}[u]$
\EndFor
\EndFor
\State $t_{\textrm{opt}} \gets \mathbf{T}\left[ \underset{u=1,...,N}{\textrm{argmax}}\frac{1}{N}\sum_{v=1}^{N}\left(\mathbf{SE}[u, v] + \mathbf{SP}[u, v]\right)\right]$
\end{algorithmic}
\end{algorithm}

\section{Experiments}
\subsection{Datasets}
\label{exp-dataset}
\begin{figure}[htp]
    \centering
    \includegraphics[width=0.9\linewidth]{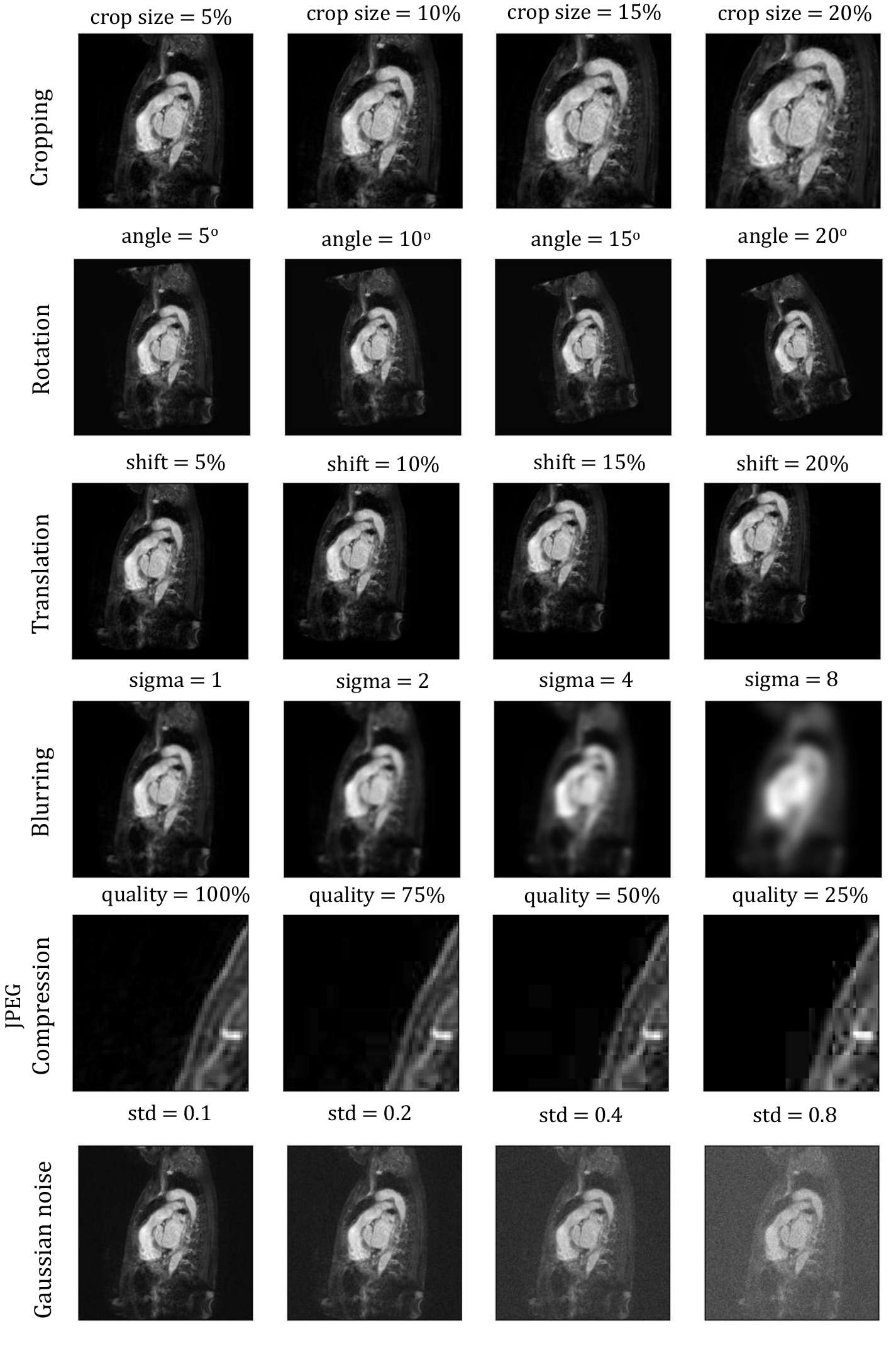}
    \caption{Near-duplicate images generated by geometric and intensity transformations under different strengths. For JPEG compression, a region of interest is shown to display the quality degradation which is not visible under human eyes when looking at the whole image.}
    \label{fig:transforms}
    \vspace{-0.05\linewidth}
\end{figure}

The Medical Segmentation Decathlon (MSD) \cite{antonelli_medical_2021} is a biomedical image analysis challenge comprising segmentation tasks related to different body organs and imaging modalities. The images are either CT or MRI of different sequences that were acquired across multiple institutions and represent real-world data. The dataset was originally split by the challenge organizer into a train and test set which consist of 1723 and 887 cases respectively. To build our benchmark, we further split the MSD train set into 3 buckets 1A, 1B, 1C, and the MSD test set into 3 buckets 2A, 2B, 2C as indicated in Table \ref{tab:data-split}. 
%We denote the train and test set by 1 and 2 and their partitions by A, B, and C. 
Buckets 1A/2A consist of the first 50\% of train/test cases across all tasks and are used to build the database and query for duplicate images. Buckets 1B/2B contain the same cases as buckets 1A/2A but with different image transformations to generate near-duplicate images. Buckets 1C/2C consist of the second 50\% of train/test cases and serve as the query for non-duplicate images. We considered each case as a unique patient and leveraged embeddings of different feature extractors, see Section \ref{sub:feature_extract}, to retrieve similar images of the query patient in the database. We used buckets 1A/1B/1C to find the decision threshold for (near-) duplicate and non-duplicate classification (see Algorithm \ref{alg:cutoff}) and evaluated this threshold on bucket 2A/2B/2C (Table \ref{tab:optimal-threshold-full-result}). Buckets 1B and 2B contain synthetic near-duplicate images by applying the following transformations with package Scipy on original images with different strengths:
\begin{itemize}
    \item Cropping: crop border in x, y, z dimensions with 5\%, 10\%, 15\%, and 20\% of the original full size
    \item Rotation: rotate the (x, y) plane by 5, 10, 15, and 20 degrees
    \item Translation: shift the images in x and y dimensions by 5\%, 10\%, 15\%, 20\% of the original size
    \item Blurring: blur images with increasing sigma of 1, 2, 4, and 8
    \item JPEG Compression: compress images under different quality of 100\%, 75\%, 50\%, and 25\%
    \item Gaussian noise: add Gaussian noise with increasing standard deviation of 0.1, 0.2, 0.4 and 0.8
\end{itemize}
The first level of strength across all transformations aims to simulate realistic preprocessing steps applied to datasets while the rest 3 levels serve as extreme cases to test the robustness of feature embeddings. Figure \ref{fig:transforms} shows examples of transformed heart MRI images.
\vspace{-0.04\linewidth}
\begin{table}[htp]
    \centering
    \caption{The number of cases per bucket}
    \label{tab:data-split}
    \begin{tabular}{c c c c}
        \toprule
        {} & Bucket 1 & Bucket 2 & Usage \\
        \midrule
        A & 859 & 443 & Database \& duplicate query \\
        B & 859 & 443 & Synthetic near-duplicate query\\
        C & 864 & 444 & Non-duplicate query\\
        %\midrule
        %Total (unique) & 1723 & 887 & {} \\
        \bottomrule
    \end{tabular}
    \vspace{-0.04\linewidth}
\end{table}
\subsection{Feature extraction and indexing}\label{sub:feature_extract}
Our selected feature extractors are DINOv1 \cite{caron2021emerging} and DINOv2 \cite{oquab2023dinov2}, which are powerful self-supervised pretrained models on natural images. The image preprocessing steps included min-max scaling of images to 0-1 and resizing the axial slices to 224$\times$224. Embeddings in buckets 1A and 2A are indexed with Locality-Sensitivity Hashing (LSH) \cite{charikar2002similarity} and Hierarchical Navigable Small World (HNSW) \cite{malkov2018efficient} for effective similarity search. 
\vspace{-0.05\linewidth}
\subsection{Evaluation}
\label{exp-evaluation}
% \vspace{-0.06\linewidth}
 We evaluated the classification in two stages. In the first stage, a query image is considered duplicate if $c(k)$ surpasses a threshold value $t$. Here we considered $c(1)$ and $c(3)$ for top-1 and top-3 retrievals, respectively. We used the Area Under the Curve of the Receiver Operating Characteristic (AUC-ROC) to evaluate the class separability at different threshold values.
 In the second stage, we took the threshold $t_{ROC}$, defined in line 5 of Algorithm \ref{alg:cutoff}, to classify the queries into positive and negative outputs. Subsequently, we compared the highest-count case ID $X^{(i_1)}$ from each case classified as duplicate to the ground-truth case IDs. If a match is found, the query output is classified as a true positive. Otherwise, it is labeled as a false positive. We computed in this stage the sensitivity and specificity values associated with $t_{ROC}$.
 
\section{Results}
\label{results}

\subsection{Decision threshold determination}
\label{results-train}
In this section, we analyzed embeddings from buckets 1A, 1B, and 1C to determine the best threshold for (near-) duplicate detection using Algorithm \ref{alg:cutoff}.
\subsubsection{Duplicate vs. Non-duplicate}
\label{train-dup-non}

\begin{figure}[htp]
    \centering
    \includegraphics[width=\linewidth]{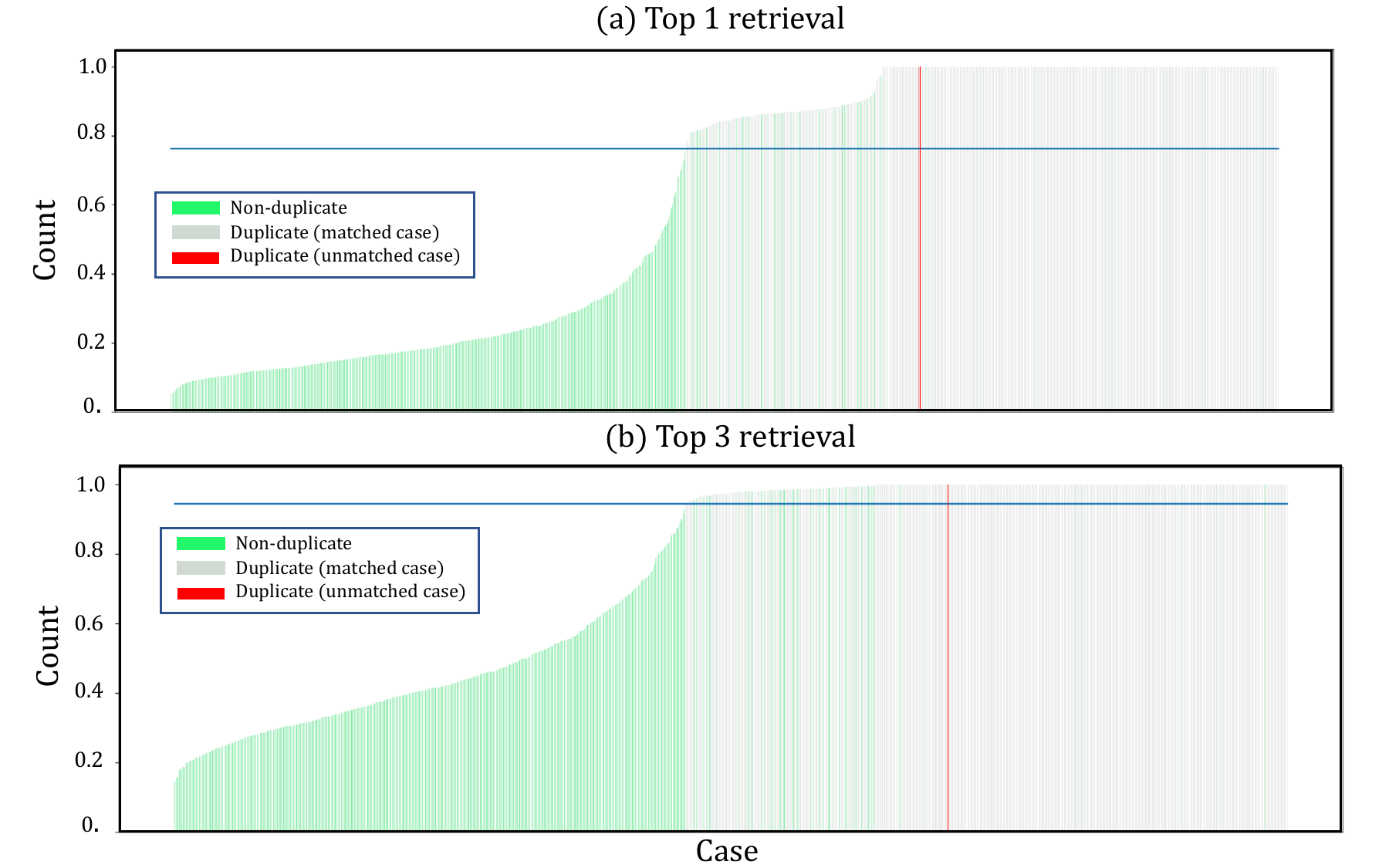}
    \caption{Case-level normalized counts of top 1 and top 3 predictions indexed with HNSW using DINOv1 embeddings. Green bars denote non-duplicate queries and gray and red bars denote duplicate queries. Red bars show duplicate queries in which the top 1 predictions are in the database but do not match the ground truth image case ID. The blue horizontal bar is the threshold resulting in the maximum sum of sensitivity and specificity.}
    \label{fig:dist_dino_v1_hnsw_top_3}
    % \vspace{-0.05\linewidth}
\end{figure}

As an overview of all thresholds, Table \ref{tab:train-metrics} displays the AUC for the first stage evaluation. The top-1 retrieval achieves AUC in the range of 0.98-0.99, while the top-3 retrievals exhibit a slight decrease to 0.96-0.98. 
Figure \ref{fig:dist_dino_v1_hnsw_top_3} illustrates the accumulated count distribution for top-1 and top-3 retrievals indexed by HNSW using DINOv1 embeddings. It is evident that extending from top-1 to top-3 retrievals results in a significant increase in counts, boosting sensitivity while decreasing specificity due to higher false positives. 
We identified $t_{ROC}$ for all top-\emph{k} retrievals and computed the associated sensitivity and specificity for the second stage (Table \ref{tab:train-metrics}). Stage 2 maintains a stable sensitivity rate of 0.99 across top-\emph{k} retrievals, with slightly lower specificity in the top 3 due to increased false positives in the first stage. However, false positives stemming from case ID mismatches in the second stage are relatively low. We also observed that HNSW indexing offers slightly higher sensitivity than LSH, while LSH provides higher specificity compared to HNSW. In terms of extractors, DINOv2 outperformed DINOv1 by a small margin.
\vspace{-0.05\linewidth}
\begin{table}[htp]
    \centering
    \caption{(Bucket 1A-1C) Duplicate detection evaluation}
    \label{tab:train-metrics}
    \begin{tabular}{c c | c c | c c}
        \toprule
         &  & \multicolumn{2}{c|}{DINOv1} & \multicolumn{2}{c}{DINOv2} \\
         &    & LSH   & HNSW  & LSH   & HNSW \\
        \midrule
        AUC & Top 1 & 0.9884 & 0.9873 & \textbf{0.9901} & 0.9898 \\
        (Stage 1) & Top 3 & 0.9658 & 0.9834 & 0.9681 & \textbf{0.9888} \\
        \midrule
        Sensitivity & Top 1 & 0.9930 & 0.9942 & \textbf{0.9965} & 0.9953 \\
        (Stage 2) & Top 3 & 0.9965 & \textbf{0.9988} & 0.9977 & \textbf{0.9988} \\
        \midrule
        Specificity & Top 1 & 0.9295 & 0.9225 & \textbf{0.9353} & 0.9318 \\
        (Stage 2) & Top 3 & 0.9306 & 0.9156 & \textbf{0.9353} & 0.9272 \\
        \bottomrule
    \end{tabular}
    \vspace{-0.05\linewidth}
\end{table}
% 
% \vspace{-0.05\linewidth}
% \begin{table}[htp]
%     \centering
%     \caption{(Stage 2 - Train) Sensitivity of duplicate detection}
%     \label{tab:sens-dup-ndup}
%     \begin{tabular}{c c c c c}
%         \toprule
%         Extractor & \multicolumn{2}{c}{DINOv1} & \multicolumn{2}{c}{DINOv2} \\
%         \midrule
%         Index   & LSH   & HNSW  & LSH   & HNSW \\
%         Top 1 & 0.9930 & 0.9942 & \textbf{0.9965} & 0.9953 \\
%         Top 3 & 0.9965 & \textbf{0.9988} & 0.9977 & \textbf{0.9988} \\
%         \midrule
%         Specificity & Top 1 & 0.9295 & 0.9225 & \textbf{0.9353} & 0.9318 \\
%         Stage 2 & Top 3 & 0.9306 & 0.9156 & \textbf{0.9353} & 0.9272 \\
%         \bottomrule
%     \end{tabular}
% \end{table}
% % 
% \vspace{-0.05\linewidth}

% \begin{table}[htp]
%     \centering
%     \caption{(Stage 2 - Train) Specificity of duplicate detection}
%     \label{tab:specs-dup-ndup}
%     \begin{tabular}{c c c c c}
%         \toprule
%         Extractor & \multicolumn{2}{c}{DINOv1} & \multicolumn{2}{c}{DINOv2} \\
%         \midrule
%         Index   & LSH   & HNSW  & LSH   & HNSW \\
%         Top 1 & 0.9295 & 0.9225 & \textbf{0.9353} & 0.9318 \\
%         Top 3 & 0.9306 & 0.9156 & \textbf{0.9353} & 0.9272 \\
%         \bottomrule
%     \end{tabular}
% \end{table}
% \vspace{-0.05\linewidth}

% 
\subsubsection{Near-duplicate vs. Non-duplicate}
\label{train-np-non-dup}
Table \ref{tab:near-dup-auc} displays the stage 1's AUC of the top 3 retrievals indexed by HNSW and using DINOv1 embeddings. The ROC curves for top 1 with combinations of other indices and feature extractors follow the same tendency and hence are not shown. 
Under lowest strength, cropped, translated, noisy, and JPEG-compressed embeddings maintained AUC scores above 0.95, followed by rotation at 0.9327 and blurring at 0.9360. When stronger transformations were considered, all transformations showed an expected decrease in AUC scores proportional to the strength.
At the strongest level, the AUC scores by blurring and Gaussian noise rose by a small margin compared to the preceded level. We postulated that under this range, the images are significantly altered and the embeddings are matched to non-informative slices of different volumes. Therefore even though the AUC scores seem to increase in stage 1, they show a consistent decrease in stage 2 where case IDs are matched (Table \ref{tab:near-dup-sens}). Regarding the feature extractors, we observed a lower performance of DINOv2 compared to DINOv1 by a margin of 0.1-0.3 AUC. Nevertheless, the best and worst transformations are consistent between the two types of embeddings.
\vspace{-0.03\linewidth}

\begin{table}[htp]
    \centering
    \caption{(Bucket 1B-1C, Stage 1) AUC of near-duplicates classification using DINOv1 embeddings with HNSW index for top 3 retrieval}
    \label{tab:near-dup-auc}
    \resizebox{\columnwidth}{!}{%
    \begin{tabular}{c | c c c c}
        \toprule
        {Transform} & \multicolumn{4}{c}{Strength of transformation}  \\
        \midrule
        Cropping   & crop=5\%  & crop=10\%  & crop=15\%   & crop=20\% \\
        {} & 0.9830 & 0.9618 & 0.8817 & 0.8024 \\
        \midrule
        Rotation & angle=5\textsuperscript{o} & angle=10\textsuperscript{o} & angle=15\textsuperscript{o} & angle=20\textsuperscript{o} \\
        {} & 0.9327 & 0.9053 & 0.8645 & 0.8426 \\
        \midrule
        Translation & shift=5\% & shift=10\% &shift=15\% &shift=20\% \\
        {} & 0.9619 & 0.9493 & 0.9388 & 0.9209 \\
        \midrule
        Blurring & sigma=1 & sigma=2 & sigma=4 & sigma=8 \\
        {} & 0.9360 & 0.7127 & 0.6916 & 0.7711 \\
        \midrule
        JPEG & quality=100\% & quality=75\% & quality=50\% & quality=25\% \\
        compression & 0.9821 & 0.9699 & 0.9405 & 0.8816 \\ 
        \midrule
        Gaussian & std=0.1 & std=0.2 &std=0.4 &std=0.8 \\
        noise & 0.9675 & 0.8437 & 0.6456 & 0.6789 \\
        \bottomrule
    \end{tabular}
    }
\vspace{-0.05\linewidth}

\end{table}
\begin{table}[htp]
    \centering
    \caption{(Bucket 1B-1C, Stage 2) Sensitivity of near-duplicates classification using DINOv1 embeddings with HNSW index for top 3 retrieval}
    \label{tab:near-dup-sens}
    \resizebox{\columnwidth}{!}{%
    \begin{tabular}{c | c c c c}
        \toprule
        {Transform} & \multicolumn{4}{c}{Strength of transformation}  \\
        \midrule
        Cropping   & crop=5\%  & crop=10\%  & crop=15\%   & crop=20\% \\
        {} & 0.9942 & 0.9604 & 0.8161 & 0.6286 \\
        \midrule
        Rotation & angle=5\textsuperscript{o} & angle=10\textsuperscript{o} & angle=15\textsuperscript{o} & angle=20\textsuperscript{o} \\
        {} & 0.8778 & 0.8522 & 0.8300 & 0.8114 \\
        \midrule
        Translation & shift=5\% & shift=10\% &shift=15\% &shift=20\% \\
        {} & 0.9581 & 0.8708 & 0.8545 & 0.8510 \\
        \midrule
        Blurring & sigma=1 & sigma=2 & sigma=4 & sigma=8 \\
        {} & 0.8196 & 0.5239 & 0.1735 & 0.0163 \\
        \midrule
        JPEG & quality=100\% & quality=75\% & quality=50\% & quality=25\% \\
        compression & 0.9709 & 0.9267 & 0.8468 & 0.8102 \\ 
        \midrule
        Gaussian & std=0.1 & std=0.2 &std=0.4 &std=0.8 \\
        noise & 0.9779 & 0.8568 & 0.3050 & 0.0664 \\
        \bottomrule
    \end{tabular}
    }
\end{table}
\subsubsection{Selected optimal threshold}
\label{results-cutoff-selection}
After observing the retrieval performance using buckets 1A, 1B, and 1C, we decided to prioritize the top 3 retrievals using the HNSW index. Even though this choice slightly decreases specificity, it offers higher sensitivity as a trade-off. We determined $t_{opt}$ on the train set (Algorithm \ref{alg:cutoff}) with 7 query sets, including 6 near-duplicate query sets of the lowest transformation strength and 1 duplicate query set. The chosen $t_{opt}$ is 0.7711 and will be used to evaluate the detection performance using buckets 2A (duplicates), 2B (near-duplicates), and 2C (non-duplicates).
\subsection{Evaluation with selected decision threshold}
\label{results-test}
In Table \ref{tab:optimal-threshold-full-result}, we present the best sensitivity and specificity results for (near-) duplicate detection on buckets 2A, 2B, 2C using the top-3 retrievals obtained through DINOv1 embeddings with HNSW indexing. Overall, for stage 1, the mean sensitivity score is 0.9645, while for stage 2, it is 0.9407. In terms of mean specificity, we found scores of 0.8559 for stage 1 and 0.8373 for stage 2. Rotation exhibits the lowest sensitivity, mirroring trends observed in the training dataset. We also observed some false positive occurrences that are indeed in our database, which is due to the fact that possible identical images are shared among different MSD tasks. Upon scanning through 10 tasks, we were able to find 28 pairs of possible (near-) duplicates in the original MSD dataset shown in Table \ref{tab:detected-duplicates}. The majority of detected (near-) duplicates are from task 1, which contains the MRI brain images. An example is shown in Figure \ref{fig:detected_duplicate} where two cases are different only in the brightness.

\begin{table}[htp!]
    \centering
    \caption{(Bucket 2A-2B-2C) Sensitivity and specificity using the threshold of 0.7711 for DINOv1 embeddings with HNSW index}
    \label{tab:optimal-threshold-full-result}
    \begin{tabular}{c | c c | c c}
        \toprule
        ~ & \multicolumn{2}{c|}{Sensitivity} & \multicolumn{2}{c}{Specificity} \\
        Transform  & Stage 1   & Stage 2  & Stage 1   & Stage 2 \\\midrule
        No transform (duplicate) & 0.9977 & 0.9977 & 0.8559 & 0.8559 \\
        Cropping (crop=5\%) & 0.9865 & 0.9865 & 0.8559 & 0.8559\\
        Translation (shift=5\%) & 0.9481 & 0.9481 & 0.8559 & 0.8559\\
        Rotation (angle=5\textsuperscript{o})& 0.8849 & 0.8826 & 0.8559 & 0.8539\\
        Blurring (sigma=1) & 0.9639 & 0.8442 & 0.8559 & 0.7646\\
        JPEG (quality=100\%) & 0.9977 & 0.9549 & 0.8559 & 0.8207\\
        Noise (std=0.1) & 0.9729 & 0.9707 & 0.8559 & 0.8539\\ 
        \midrule
        Mean & 0.9645 & 0.9407 & 0.8559 & 0.8373\\
        \bottomrule
    \end{tabular}
    % \vspace{-0.03\linewidth}
\end{table}

\section{Conclusion}
In this study, we introduced and evaluated an unsupervised approach for near- and duplicate detection of 3D medical images relying on pretrained 2D vision embeddings. We quantitatively assessed its performance to detect near-duplicates involving various image transformations and different strength levels. A heuristic based on Youden's index is proposed to select a single threshold value for balanced performance in the (near-) duplicate detection task. Our top results show 0.9645 sensitivity and 0.8559 specificity for detecting duplicates and near-duplicates. This underscores the transferability of pretrained embeddings from natural images to medical imaging. Additionally, our method successfully pinpointed potential (near-) duplicates in MSD datasets, validating its effectiveness in identifying real-world duplicates. We consider extending our study with in-depth analysis under different datasets, modalities, contrasts, and organs, as well as investigating naturally occurring near duplicates, e.g., same patient at different time points, together with different volume sampling strategies to address spatial redundancy in future work.

\bibliographystyle{IEEEtran}
\bibliography{ref}
\section*{Compliance with Ethical Standards}

This research study was conducted retrospectively using human subject data made available in open access by \url{https://medicaldecathlon.com/}. Ethical approval was not required as confirmed by the license attached with the open access data.

\section*{Conflicts of Interest}
No funding was received for conducting this study. The authors have no relevant financial or non-financial interests to disclose.

\newpage
\onecolumn

\section*{Appendix}
\renewcommand{\thetable}{\Alph{subsection}.\arabic{table}} % Set table numbering format to "A.#"
\renewcommand{\thefigure}{\Alph{subsection}.\arabic{figure}} % Set table numbering format to "A.#"
\renewcommand{\theHsection}{A\arabic{subsection}}
% \counterwithin{figure}{section}
% \counterwithin{table}{section}
\setcounter{table}{0}

\subsection{Near-duplicate detection results}
\label{appendix:near-duplicate-results}
We present in this section the full results of near-duplicate analysis on buckets 1A, 1B, and 1C. The evaluation was conducted using two feature extractors DINOv1 and DINOv2, and two database indices LSH and HNSW. For stage 1 (Table \ref{tab:full-auc-near-dup-results}), we report only AUC, and for stage 2, we report the sensitivity (Table \ref{tab:full-sens-near-dup-results}) and specificity (Table \ref{tab:full-specs-near-dup-results}) associated with $t_{ROC}$. For all metrics, we show the results with top-1 and top-3 retrieval.
% AUC of top 1
% \vspace{-0.05\linewidth}
\begin{table}[htp]
    \centering
    \caption{(Bucket 1B-1C, Stage 1) AUC of near-duplicates classification using DINOv1 and DINOv2 embeddings with LSH and HNSW indexing. \textbf{Left}: top-1 retrieval. \textbf{Right}: top-3 retrieval}
    \label{tab:full-auc-near-dup-results}
    % \resizebox{\columnwidth}{!}{%
        \begin{tabular}[t]{c  c  c c  c c}
            \toprule
            \multirow{2}{*}{Transform} &   \multirow{2}{*}{Strength} & \multicolumn{2}{c}{DINOv1} & \multicolumn{2}{c}{DINOv2}\\
            \cmidrule{3-6}
            & {} & LSH & HNSW & LSH & HNSW \\
            \midrule
            \multirow{4}{*}{Crop} & crop=5\% & \textbf{0.9902} & 0.9887 & 0.9661 & 0.9732 \\
                                  & crop=10\% & 0.9751 & \textbf{0.9762} & 0.8964 & 0.9102 \\
                                  & crop=15\% & 0.8736 & \textbf{0.8932} & 0.7624 & 0.7835 \\ 
                                  & crop=20\% & 0.7876 & \textbf{0.7989} & 0.7190 & 0.7193 \\
            \midrule
            \multirow{4}{*}{Rotation} 
            & angle=5\textsuperscript{o} & 0.9428 & \textbf{0.9457} & 0.9421 & 0.9461 \\ 
            & angle=10\textsuperscript{o} & 0.9156 & \textbf{0.9202} & 0.8864 & 0.8864 \\
            & angle=15\textsuperscript{o} & \textbf{0.8791} & 0.8775 & 0.8092 & 0.7912 \\
            & angle=20\textsuperscript{o} & 0.8576 & \textbf{0.8621} & 0.7225 & 0.6874 \\
            \midrule
            \multirow{4}{*}{Translation} 
            & shift=5\% & \textbf{0.9688} & 0.9678 & 0.9675 & 0.9672 \\
            & shift=10\% & 0.9578 & 0.9577 & \textbf{0.9587} & 0.9580 \\
            & shift=15\% & 0.9474 & 0.9476 & \textbf{0.9482} & 0.9447 \\ 
            & shift=20\% & 0.9270 & 0.9290 & \textbf{0.9327} & 0.9273 \\
            \midrule
            \multirow{4}{*}{Blurring}
            & sigma=1 & 0.9385 & \textbf{0.9437} & 0.9008 & 0.8940 \\ 
            & sigma=2 & 0.7031 & \textbf{0.7164} & 0.6774 & 0.6529 \\ 
            & sigma=4 & 0.6330 & 0.6778 & \textbf{0.7456} & 0.6848 \\
            & sigma=8 & 0.7043 & 0.7534 & \textbf{0.8232} & 0.7470 \\
            \midrule
            {}          & quality=100\% & \textbf{0.9820} & 0.9810 & 0.9830 & 0.9811 \\ 
            JPEG        & quality=75\% & 0.9736 & \textbf{0.9741} & 0.9636 & 0.9618 \\
            Compression & quality=50\% & 0.9480 & \textbf{0.9504} & 0.9344 & 0.9247 \\ 
            {}          & quality=25\% & 0.8745 & \textbf{0.8894} & 0.8422 & 0.8562 \\
            \midrule
            {}          & std=0.1 & 0.9621 & \textbf{0.9681} & 0.9292 & 0.9380 \\
            Gaussian    & std=0.2 & 0.8195 & \textbf{0.8485} & 0.7956 & 0.8020 \\
            noise       & std=0.4 & 0.6069 & \textbf{0.6354} & 0.5976 & 0.5951 \\ 
            {}          & std=0.8 & 0.6451 & \textbf{0.6571} & 0.6114 & 0.6248 \\
            \bottomrule
        \end{tabular}
        \begin{tabular}[t]{c  c  c c  c c}
            % \caption{Top-3 retrieval}
            \toprule
            \multirow{2}{*}{Transform} &   \multirow{2}{*}{Strength} & \multicolumn{2}{c}{DINOv1} & \multicolumn{2}{c}{DINOv2}\\
            \cmidrule{3-6}
            & {} & LSH & HNSW & LSH & HNSW \\
            \midrule
            \multirow{4}{*}{Crop} & crop=5\%  & 0.9579 & \textbf{0.9830} & 0.9351 & 0.9581 \\
                                  & crop=10\% & 0.9372 & \textbf{0.9618} & 0.8661 & 0.8807 \\
                                  & crop=15\% & 0.8528 & \textbf{0.8817} & 0.7622 & 0.7781 \\ 
                                  & crop=20\% & 0.7906 & \textbf{0.8024} & 0.7276 & 0.7282 \\
            \midrule
            \multirow{4}{*}{Rotation} 
            & angle=5\textsuperscript{o}  & 0.9579 & \textbf{0.9830} & 0.9351 & 0.9581 \\ 
            & angle=10\textsuperscript{o} & 0.9372 & \textbf{0.9618} & 0.8661 & 0.8807 \\
            & angle=15\textsuperscript{o} & 0.8528 & \textbf{0.8817} & 0.7622 & 0.7781 \\
            & angle=20\textsuperscript{o} & 0.7906 & \textbf{0.8024} & 0.7276 & 0.7282\\
            \midrule
            \multirow{4}{*}{Translation} 
            & shift=5\%  & 0.9411 & \textbf{0.9619} & 0.9399 & 0.9593 \\
            & shift=10\% & 0.9278 & \textbf{0.9493} & 0.9260 & 0.9448\\
            & shift=15\% & 0.9185 & \textbf{0.9388} & 0.9128 & 0.9293\\ 
            & shift=20\% & 0.9013 & \textbf{0.9209} & 0.8970 & 0.9124\\
            \midrule
            \multirow{4}{*}{Blurring}
            & sigma=1 &  0.9164 & \textbf{0.9360} & 0.8811 & 0.8817\\ 
            & sigma=2 &  0.7005 & \textbf{0.7127} & 0.7013 & 0.6773\\ 
            & sigma=4 &  0.6561 & 0.6916 & \textbf{0.7752} & 0.7267\\
            & sigma=8 &  0.7191 & 0.7711 & \textbf{0.8515} & 0.7902\\
            \midrule
            {}          & quality=100\% & 0.9637 & 0.9821 & 0.9639 & \textbf{0.9841}\\ 
            JPEG        & quality=75\%  & 0.9474 & \textbf{0.9699} & 0.9385 & 0.9535\\
            Compression & quality=50\%  & 0.9185 & \textbf{0.9405} & 0.9038 & 0.9141\\ 
            {}          & quality=25\%  & 0.8531 & \textbf{0.8816} & 0.8213 & 0.8328\\
            \midrule
            {}          & std=0.1 & 0.9329 & \textbf{0.9675} & 0.9015 & 0.9090\\
            Gaussian    & std=0.2 & 0.8002 & \textbf{0.8437} & 0.7696 & 0.7618\\
            noise       & std=0.4 & 0.6217 & \textbf{0.6456} & 0.6104 & 0.5814\\ 
            {}          & std=0.8 & 0.6573 & \textbf{0.6789} & 0.6296 & 0.6190\\
            \bottomrule
        \end{tabular}
    % }
\end{table}
\begin{table}[htp]
    \centering
    \caption{(Bucket 1B-1C, Stage 2) Sensitivity of near-duplicates classification using DINOv1 and DINOv2 embeddings with LSH and HNSW indexing. \textbf{Left}: top-1 retrieval. \textbf{Right}: top-3 retrieval}
    \label{tab:full-sens-near-dup-results}
    \begin{tabular}[t]{c  c  c c  c c}
        \toprule
        \multirow{2}{*}{Transform} &   \multirow{2}{*}{Strength} & \multicolumn{2}{c}{DINOv1} & \multicolumn{2}{c}{DINOv2}\\
        \cmidrule{3-6}
        & {} & LSH & HNSW & LSH & HNSW \\
        \midrule
        \multirow{4}{*}{Crop} & crop=5\% &  \textbf{0.9732} & 0.9208 & 0.9627 & 0.6321\\
                              & crop=10\% & 0.9464 & \textbf{0.9546} & 0.6764 & 0.7229 \\
                              & crop=15\% & \textbf{0.7695} & 0.7637 & 0.4517 & 0.4319 \\ 
                              & crop=20\% &  0.4843 & \textbf{0.5204} & 0.3504 & 0.3667\\
        \midrule
        \multirow{4}{*}{Rotation} 
        & angle=5\textsuperscript{o} &  0.8952 & \textbf{0.9371} & 0.9313 & 0.9348\\ 
        & angle=10\textsuperscript{o} & \textbf{0.8359 }& 0.8044 & 0.7416 & 0.5460\\
        & angle=15\textsuperscript{o} & \textbf{0.7811} & 0.7765 & 0.2992 & 0.2794 \\
        & angle=20\textsuperscript{o} & \textbf{0.7625} & 0.7590 & 0.1746 & 0.1513\\
        \midrule
        \multirow{4}{*}{Translation} 
        & shift=5\% & 0.9488 & \textbf{0.9756} & 0.9418 & 0.9686\\
        & shift=10\% & 0.9034 & \textbf{0.9395} & 0.8917 & 0.9139\\
        & shift=15\% & 0.8685 & \textbf{0.9127} & 0.8417 & 0.8533\\ 
        & shift=20\% & 0.8405 & \textbf{0.8591} & 0.8231 & 0.8300\\
        \midrule
        \multirow{4}{*}{Blurring}
        & sigma=1 &  \textbf{0.7998} & 0.7974 & 0.7509 & 0.7683\\ 
        & sigma=2 &  0.3760 & \textbf{0.4144} & 0.1735 & 0.1921\\ 
        & sigma=4 &  0.0861 & \textbf{0.3539} & 0.0442 & 0.0373\\
        & sigma=8 & 0.0093 & 0.0035 & \textbf{0.0116} & 0.0023\\
        \midrule
        {}          & quality=100\% &  0.9383 & \textbf{0.9464} & 0.9302 & 0.9360\\ 
        JPEG        & quality=75\% &  \textbf{0.9302} & 0.8941 & 0.8766 & 0.8091\\
        Compression & quality=50\% &  0.8778 & \textbf{0.8941} & 0.8172 & 0.8149\\ 
        {}          & quality=25\% & 0.7870 & \textbf{0.8056} & 0.6973 & 0.7055\\
        \midrule
        {}          & std=0.1 &  \textbf{0.9616} & 0.9325 & 0.8917 & 0.7322\\
        Gaussian    & std=0.2 &  0.7113 & \textbf{0.7474} & 0.6228 & 0.6403\\
        noise       & std=0.4 &  0.2701 & \textbf{0.2864} & 0.2375 & 0.2177\\
        {}          & std=0.8 &  0.0268 & 0.0326 & \textbf{0.0384} & 0.0349\\
        \bottomrule
    \end{tabular}
    % }
   \begin{tabular}[t]{c  c  c c  c c}
        \toprule
        \multirow{2}{*}{Transform} &   \multirow{2}{*}{Strength} & \multicolumn{2}{c}{DINOv1} & \multicolumn{2}{c}{DINOv2}\\
        \cmidrule{3-6}
        & {} & LSH & HNSW & LSH & HNSW \\
        \midrule
        \multirow{4}{*}{Crop} & crop=5\% &  \textbf{0.9942} & \textbf{0.9942} & 0.7322 & 0.7288\\
                              & crop=10\% & 0.9325 & \textbf{0.9604} & 0.6310 & 0.7159\\
                              & crop=15\% & 0.7800 & \textbf{0.8161} & 0.3935 & 0.3970\\ 
                              & crop=20\% & 0.5972 & \textbf{0.6286} & 0.3411 & 0.3283\\
        \midrule
        \multirow{4}{*}{Rotation} 
        & angle=5\textsuperscript{o} &  0.8731 & \textbf{0.8778} & 0.8417 & 0.8359\\ 
        & angle=10\textsuperscript{o} & 0.8428 & \textbf{0.8522} & 0.5274 & 0.5320\\
        & angle=15\textsuperscript{o} & 0.8161 & \textbf{0.8300} & 0.3143 & 0.3399\\
        & angle=20\textsuperscript{o} & 0.8068 & \textbf{0.8114} & 0.2456 & 0.2212\\
        \midrule
        \multirow{4}{*}{Translation} 
        & shift=5\% & 0.9430 & \textbf{0.9581} & 0.8987 & 0.8941\\
        & shift=10\% & \textbf{0.9010} & 0.8708 & 0.8568 & 0.8312 \\
        & shift=15\% & 0.8487 & \textbf{0.8545}  & 0.7497 & 0.7718\\ 
        & shift=20\% & 0.8428 & \textbf{0.8510 }& 0.7485 & 0.8126\\
        \midrule
        \multirow{4}{*}{Blurring}
        & sigma=1 &  0.8091 & \textbf{0.8196} & 0.6938 & 0.6868\\ 
        & sigma=2 &  \textbf{0.5704} & 0.5239 & 0.3120 & 0.2305\\ 
        & sigma=4 &  0.1538 & \textbf{0.1735} & 0.0675 & 0.0477\\
        & sigma=8 & \textbf{0.0256} & 0.0163 & 0.0221 & 0.0128\\
        \midrule
        {}          & quality=100\% & 0.9686 & \textbf{0.9709} & 0.9197 & 0.9616\\ 
        JPEG        & quality=75\% & \textbf{0.9499} & 0.9267 & 0.8789 & 0.8533\\
        Compression & quality=50\% & 0.8440 & \textbf{0.8498} & 0.6985 & 0.6612\\ 
        {}          & quality=25\% & 0.7858 & \textbf{0.8102} & 0.5844 & 0.5390\\
        \midrule
        {}          & std=0.1 &  \textbf{0.9837} & 0.9779 & 0.7811 & 0.4924\\
        Gaussian    & std=0.2 &  0.8207 & \textbf{0.8568 }& 0.6286 & 0.4889\\
        noise       & std=0.4 &  0.2992 & \textbf{0.3050} & 0.2398 & 0.1665\\
        {}          & std=0.8 &  0.0640 & \textbf{0.0664} & 0.0629 & 0.0419\\
        \bottomrule
    \end{tabular}
\end{table}
\begin{table}[htp!]
    \centering
    \caption{(Bucket 1B-1C, Stage 2) Specificity of near-duplicates classification using DINOv1 and DINOv2 embeddings with LSH and HNSW indexing. \textbf{Left}: top-1 retrieval. \textbf{Right}: top-3 retrieval.}
    \label{tab:full-specs-near-dup-results}
    % \resizebox{\columnwidth}{!}{%
    \begin{tabular}[t]{c  c  c c  c c}
        \toprule
        \multirow{2}{*}{Transform} &   \multirow{2}{*}{Strength} & \multicolumn{2}{c}{DINOv1} & \multicolumn{2}{c}{DINOv2}\\
        \cmidrule{3-6}
        & {} & LSH & HNSW & LSH & HNSW \\
        \midrule
        \multirow{4}{*}{Crop} & crop=5\% &  0.8679 & \textbf{0.9734} & 0.8885 & 0.9746\\
                              & crop=10\% & 0.8602 & 0.8467 & \textbf{0.8848} & 0.8693\\
                              & crop=15\% & 0.7659 & 0.8344 & 0.7991 & \textbf{0.8731}\\ 
                              & crop=20\% & 0.4805 & 0.5306 & 0.6083 & \textbf{0.6434}\\
        \midrule
        \multirow{4}{*}{Rotation} 
        & angle=5\textsuperscript{o} &  0.8061 & 0.7768 & \textbf{0.8352} & 0.8185\\ 
        & angle=10\textsuperscript{o} & 0.8231 & 0.8925 & 0.8609 & \textbf{0.9109}\\
        & angle=15\textsuperscript{o} & 0.8754 & 0.8776 & 0.8851 & \textbf{0.8963}\\
        & angle=20\textsuperscript{o} & 0.8385 & 0.8441 & 0.8503 & \textbf{0.8534}\\
        \midrule
        \multirow{4}{*}{Translation} 
        & shift=5\% & 0.8667 & 0.8345 & \textbf{0.8883} & 0.8603\\
        & shift=10\% & 0.8576 & 0.8343 & \textbf{0.8730} & 0.8554\\
        & shift=15\% & 0.8567 & 0.8364 & \textbf{0.8730} & 0.8534\\ 
        & shift=20\% & 0.8557 & 0.8307 & \textbf{0.8720} & 0.8573\\
        \midrule
        \multirow{4}{*}{Blurring}
        & sigma=1 & 0.7265 & 0.7072 & \textbf{0.8503} & 0.8486\\ 
        & sigma=2 & 0.6660 & 0.6211 & \textbf{0.7151} & 0.6647\\ 
        & sigma=4 & 0.3539 & 0.2916 & \textbf{0.4315} & 0.4114\\
        & sigma=8 & 0.3241 & 0.3453 & 0.3725 & \textbf{0.4103}\\
        \midrule
        {}          & quality=100\% & 0.8626 & 0.8344 & \textbf{0.8846} & 0.8613\\ 
        JPEG        & quality=75\% & 0.8525 & 0.8931 & 0.8814 & \textbf{0.9091}\\
        Compression & quality=50\% & 0.8348 & 0.8077 & \textbf{0.8642} & 0.8429\\ 
        {}          & quality=25\% & 0.7757 & 0.7598 & \textbf{0.8203} & 0.8190\\
        \midrule
        {}          & std=0.1 & 0.8534 & 0.9087 & 0.8834 & \textbf{0.9168}\\
        Gaussian    & std=0.2 & 0.6831 & 0.6639 & \textbf{0.7895} & 0.7695\\
        noise       & std=0.4 & 0.2459 & 0.3225 & 0.4099 & \textbf{0.4854}\\
        {}          & std=0.8 & 0.3956 & 0.3979 & 0.5412 & \textbf{0.5574}\\
        \bottomrule
    \end{tabular}
    % }
    \begin{tabular}[t]{c  c  c c  c c}
        \toprule
        \multirow{2}{*}{Transform} &   \multirow{2}{*}{Strength} & \multicolumn{2}{c}{DINOv1} & \multicolumn{2}{c}{DINOv2}\\
        \cmidrule{3-6}
        & {} & LSH & HNSW & LSH & HNSW \\
        \midrule
        \multirow{4}{*}{Crop} & crop=5\% &  0.9133 & 0.9133 & 0.9260 & \textbf{0.9272}\\
                              & crop=10\% & 0.8694 & 0.8451 & \textbf{0.8798} & 0.8707\\
                              & crop=15\% & 0.8326 & 0.8466 & 0.8527 & \textbf{0.8711}\\ 
                              & crop=20\% & 0.5804 & 0.6254 & 0.7112 & \textbf{0.7180}\\
        \midrule
        \multirow{4}{*}{Rotation} 
        & angle=5\textsuperscript{o} &  0.8705 & 0.8590 & 0.8798 & \textbf{0.8890}\\ 
        & angle=10\textsuperscript{o} & 0.8555 & 0.8497 & 0.8738 & \textbf{0.8773}\\
        & angle=15\textsuperscript{o} & 0.8197 & 0.7968 & \textbf{0.8566} & 0.8472\\
        & angle=20\textsuperscript{o} & 0.7569 & 0.7344 & \textbf{0.8121} & 0.7968\\
        \midrule
        \multirow{4}{*}{Translation} 
        & shift=5\% & 0.8705 & 0.8532 & \textbf{0.8809} & \textbf{0.8809}\\
        & shift=10\% & 0.8705 & 0.9017 & 0.8798 & \textbf{0.9121}\\
        & shift=15\% & 0.9168 & 0.9133 & \textbf{0.9295} & 0.9272\\ 
        & shift=20\% & 0.9179 & 0.9110 & \textbf{0.9306} & 0.9249\\
        \midrule
        \multirow{4}{*}{Blurring}
        & sigma=1 & 0.7351 & 0.7330 & 0.8611 & \textbf{0.8661}\\ 
        & sigma=2 & 0.4257 & 0.4990 & 0.5181 & \textbf{0.5772}\\ 
        & sigma=4 & 0.2776 & 0.3527 & 0.3580 & \textbf{0.4696}\\
        & sigma=8 & 0.2775 & 0.3183 & 0.3502 & \textbf{0.3959}\\
        \midrule
        {}          & quality=100\% & 0.9211 & 0.9000 & \textbf{0.9256} & 0.9188\\ 
        JPEG        & quality=75\% & 0.8459 & 0.8740 & 0.8609 & \textbf{0.8915}\\
        Compression & quality=50\% & 0.8840 & 0.8819 & 0.8971 & \textbf{0.8982}\\ 
        {}          & quality=25\% & 0.8083 & 0.8090 & \textbf{0.8412 }& 0.8393\\
        \midrule
        {}          & std=0.1 & 0.8705 & 0.8971 & 0.8798 & \textbf{0.9109}\\
        Gaussian    & std=0.2 & 0.6626 & 0.7050 & 0.7762 & \textbf{0.7910}\\
        noise       & std=0.4 & 0.3152 & 0.3764 & 0.5063 & \textbf{0.5726}\\
        {}          & std=0.8 & 0.4276 & 0.4423 & 0.5961 & \textbf{0.6657}\\
        \bottomrule
    \end{tabular}
\end{table}
\newpage
\subsection{Detected (near-) duplicates in MSD dataset}
We scanned through the MSD datasets using our proposed approach to detect near- and duplicates across tasks and train/test subsets. We used the DINOv1 embeddings indexed with HNSW and a threshold of 0.8. Any potential (near-) duplicate flagged by the algorithm was manually reviewed and confirmed. We presented in the tables below the case IDs that were flagged as (near-) duplicates and the subset to which they belong. Case ID 1 and Case ID 2 refer to the filenames of duplicate pairs in the dataset.
\setcounter{table}{0}

\begin{table}[htp!]
    \centering
    \caption{Detected (near-) duplicates in MSD dataset}
    \label{tab:detected-duplicates}
    % \resizebox{0.25\columnwidth}{!}{%
    \begin{tabular}[t]{c c | c c }
        \toprule
        Case ID 1 & Subset & Case ID 2 & Subset\\
        \midrule
        BRATS\_432 & Train & BRATS\_142 & Train\\
        BRATS\_416 & Train & BRATS\_099 & Train\\
        BRATS\_442 & Train & BRATS\_166 & Train\\
        BRATS\_398 & Train & BRATS\_063 & Train \\
        BRATS\_082 & Train & BRATS\_404 & Train \\
        BRATS\_460 & Train & BRATS\_225 & Train \\
        BRATS\_065 & Train & BRATS\_400 & Train \\
        BRATS\_406 & Train & BRATS\_084 & Train \\
        BRATS\_445 & Train & BRATS\_172 & Train \\
        BRATS\_464 & Train & BRATS\_235 & Train \\
        BRATS\_396 & Train & BRATS\_051 & Train \\
        BRATS\_224 & Train & BRATS\_459 & Train \\
        BRATS\_457 & Train & BRATS\_220 & Train\\
        BRATS\_405 & Train & BRATS\_083 & Train \\
        BRATS\_448 & Train & BRATS\_181 & Train \\
        BRATS\_261 & Train & BRATS\_471 & Train \\
        BRATS\_550 & Train & BRATS\_549 & Train \\
        BRATS\_666 & Train & BRATS\_663 & Train \\
        BRATS\_508 & Train & BRATS\_509 & Train \\
        BRATS\_522 & Train & BRATS\_521 & Train \\
        BRATS\_717 & Test & BRATS\_226 & Train  \\
        BRATS\_691 & Test & BRATS\_052 & Train  \\
        BRATS\_720 & Test & BRATS\_243 & Train  \\
        BRATS\_710 & Test & BRATS\_169 & Train  \\
        BRATS\_719 & Test & BRATS\_233 & Train  \\
        BRATS\_685 & Test & BRATS\_030 & Train  \\
        BRATS\_721 & Test & BRATS\_253 & Train  \\
        BRATS\_687 & Test & BRATS\_032 & Train  \\
        BRATS\_727 & Test & BRATS\_270 & Train  \\
        BRATS\_683 & Test & BRATS\_022 & Train  \\
        BRATS\_701 & Test & BRATS\_130 & Train  \\
        BRATS\_690 & Test & BRATS\_047 & Train  \\
        BRATS\_715 & Test & BRATS\_221 & Train  \\
        BRATS\_726 & Test & BRATS\_268 & Train  \\
        BRATS\_702 & Test & BRATS\_135 & Train  \\
        BRATS\_725 & Test & BRATS\_262 & Train  \\
        BRATS\_711 & Test & BRATS\_170 & Train  \\
        BRATS\_689 & Test & BRATS\_042 & Train  \\
        BRATS\_693 & Test & BRATS\_077 & Train  \\
        BRATS\_698 & Test & BRATS\_117 & Train  \\
        BRATS\_684 & Test & BRATS\_025 & Train  \\ 
        BRATS\_712 & Test & BRATS\_194 & Train  \\
        BRATS\_724 & Test & BRATS\_259 & Train  \\
        BRATS\_716 & Test & BRATS\_223 & Train  \\

        \bottomrule
    \end{tabular}
    \quad
    \begin{tabular}[t]{c c | c c}
        \toprule
        Case ID 1 & Subset & Case ID 2 & Subset\\
        \midrule
        BRATS\_676 & Test & BRATS\_021 & Train  \\
        BRATS\_707 & Test & BRATS\_153 & Train  \\
        BRATS\_697 & Test & BRATS\_116 & Train  \\
        BRATS\_728 & Test & BRATS\_272 & Train  \\
        BRATS\_677 & Test & BRATS\_024 & Train  \\
        BRATS\_699 & Test & BRATS\_126 & Train  \\
        BRATS\_713 & Test & BRATS\_198 & Train  \\
        BRATS\_723 & Test & BRATS\_257 & Train  \\
        BRATS\_709 & Test & BRATS\_163 & Train  \\ 
        BRATS\_678 & Test & BRATS\_023 & Train  \\ 
        BRATS\_688 & Test & BRATS\_034 & Train  \\ 
        BRATS\_708 & Test & BRATS\_161 & Train  \\ 
        BRATS\_682 & Test & BRATS\_027 & Train  \\ 
        BRATS\_694 & Test & BRATS\_092 & Train  \\ 
        BRATS\_686 & Test & BRATS\_031 & Train  \\ 
        BRATS\_695 & Test & BRATS\_112 & Train  \\ 
        BRATS\_700 & Test & BRATS\_128 & Train  \\ 
        BRATS\_704 & Test & BRATS\_144 & Train  \\ 
        BRATS\_681 & Test & BRATS\_028 & Train  \\ 
        BRATS\_714 & Test & BRATS\_218 & Train  \\ 
        BRATS\_722 & Test & BRATS\_254 & Train  \\ 
        BRATS\_692 & Test & BRATS\_076 & Train  \\ 
        BRATS\_705 & Test & BRATS\_146 & Train  \\ 
        BRATS\_703 & Test & BRATS\_143 & Train  \\ 
        BRATS\_679 & Test & BRATS\_029 & Train  \\ 
        BRATS\_696 & Test & BRATS\_115 & Train  \\ 
        BRATS\_718 & Test & BRATS\_230 & Train  \\ 
        liver\_14 & Train & liver\_15 & Train \\
        liver\_0 & Train & liver\_1 & Train \\
        liver\_171 & Train & liver\_172 & Train \\
        liver\_169 & Train & liver\_170 & Train \\
        liver\_137 & Test & liver\_74 & Train \\
        hepaticvessel\_265 & Train & spleen\_19 & Train\\
        hepaticvessel\_286 & Train & hepaticvessel\_287 & Train \\
        hepaticvessel\_376 & Test & spleen\_18 & Train \\
        hepaticvessel\_045 & Test & hepaticvessel\_139 & Train \\
        spleen\_1 & Test & hepaticvessel\_433 & Train \\
        spleen\_32 & Train & colon\_041  & Train\\
        spleen\_48 & Train & colon\_055 & Train\\
        spleen\_7 & Test & hepaticvessel\_362 & Train \\
        pancreas\_027  & Test & pancreas\_032 & Train \\
        pancreas\_007 & Test & pancreas\_228 & Train \\
        colon\_067 & Test & spleen\_59 & Train \\
        colon\_071 & Test & spleen\_60 & Train \\
        \bottomrule
    \end{tabular}
    % }
\end{table}

\setcounter{figure}{0}
\begin{figure}[htp]
    \centering
    \includegraphics[width=\linewidth]{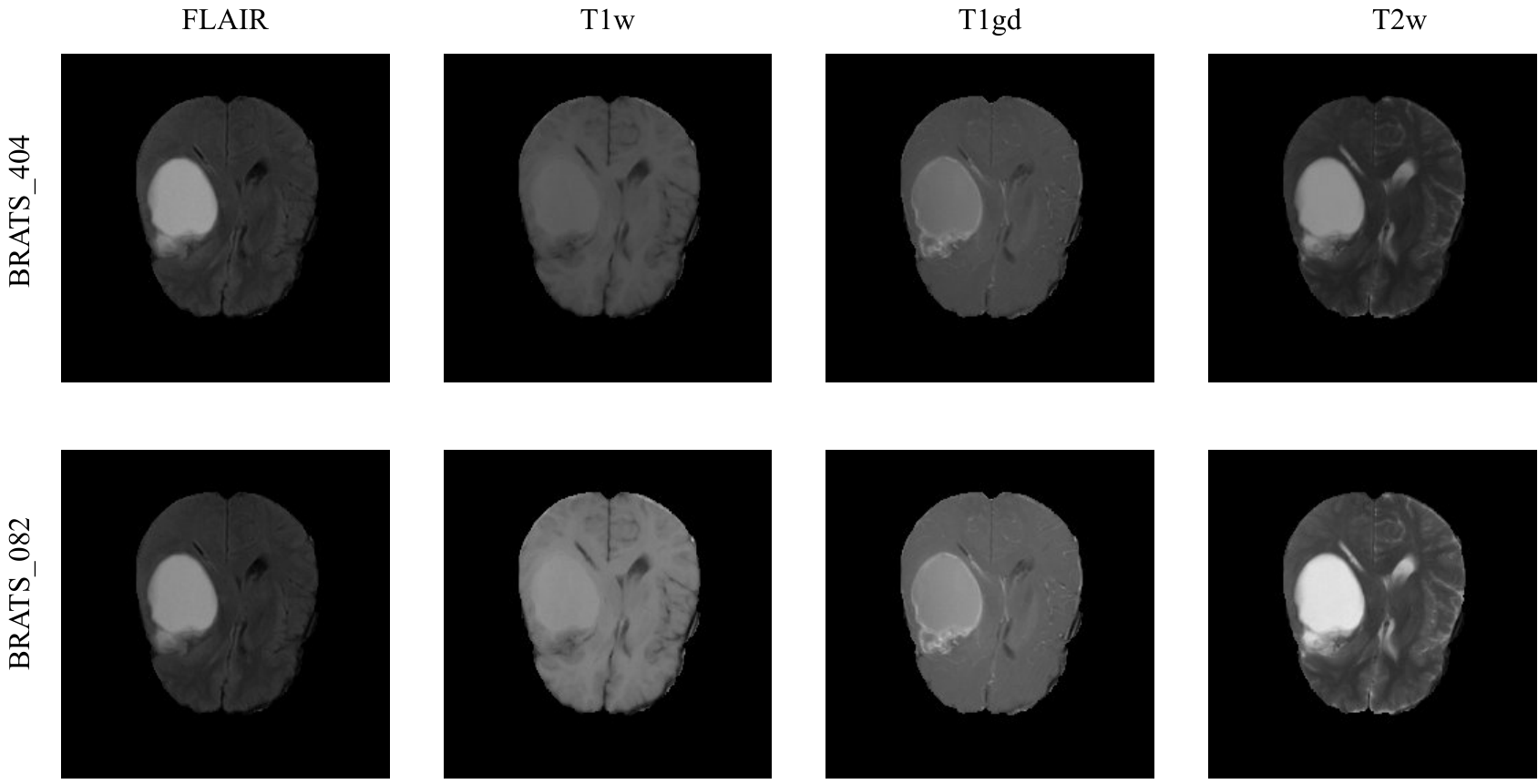}
    \caption{Example of a near-duplicate found in MSD dataset. The BRATS\_404 and BRATS\_082 are a near-duplicate pair that is different only in the brightness.}
    \label{fig:detected_duplicate}
    % \vspace{-0.05\linewidth}
\end{figure}
\end{document}